\documentclass[10pt,twocolumn,letterpaper]{article}

\usepackage{cvpr}
\usepackage{times}
\usepackage{epsfig}
\usepackage{graphicx}
\usepackage{amsmath}
\usepackage{amssymb}
\usepackage{algorithm}
\usepackage{algpseudocode}
\usepackage{threeparttable}
\usepackage{subfigure}
\usepackage{booktabs}
\usepackage{caption}

\usepackage{authblk}

\usepackage[breaklinks=true,bookmarks=false]{hyperref}

\cvprfinalcopy 

\def\cvprPaperID{****} 

\begin{document}
	
	\title{Unsupervised Reinforcement Learning of Transferable Meta-Skills \\for Embodied Navigation}
	
	\author{
		\centerline{Juncheng Li$^1$\quad Xin Wang$^2$\quad Siliang Tang$^1$ \thanks{Siliang Tang is the correspondence author.} \quad Haizhou Shi$^1$\quad Fei Wu$^1$}
		\centerline{Yueting Zhuang$^1$\quad William Yang Wang$^2$}
		\centerline{$^1$Zhejiang University\quad $^2$University of California, Santa Barbara}
		\centerline{{\tt\small {\{junchengli, siliang, shihaizhou, wufei, yzhuang\}}@zju.edu.cn , {\{xwang, william\}}@cs.ucsb.edu}}
	}
	
	\maketitle
	\pagestyle{empty}
	\thispagestyle{empty}
	
	\begin{abstract}
		Visual navigation is a task of training an embodied agent by intelligently navigating to a target object (\eg, television) using only visual observations. A key challenge for current deep reinforcement learning models lies in the requirements for a large amount of training data. It is exceedingly expensive to construct sufficient 3D synthetic environments annotated with the target object information. In this paper, we focus on visual navigation in the low-resource setting, where we have only a few training environments annotated with object information. We propose a novel unsupervised reinforcement learning approach to learn transferable meta-skills (\eg, bypass obstacles, go straight) from unannotated environments without any supervisory signals. The agent can then fast adapt to visual navigation through learning a high-level master policy to combine these meta-skills, when the visual-navigation-specified reward is provided. Experimental results show that our method significantly outperforms the baseline by $53.34\%$ relatively on SPL, and further qualitative analysis demonstrates that our method learns transferable motor primitives for visual navigation.
	\end{abstract}

	\section{Introduction}\label{s1}
	Visual navigation is a task of training an embodied agent that can intelligently navigate to an instance of an object according to the natural-language name of the object. In addition to being a fundamental scientific goal in computer vision and artificial intelligence, navigation in a 3D environment is a crucial skill for the embodied agent. This task may benefit many practical applications where an embodied agent improves the quality of life and augments human capability, such as in-home robots, personal assistants, and hazard removal robots. 
	
	Recently, various deep reinforcement learning~(DRL) approaches~\cite{wortsman2019learning, mousavian2018visual, wang2018look, wang2019reinforced, savva2017minos, wu2018building, yu2018guided,gupta2017cognitive, mirowski2016learning, zhu2017target, li2019walking} have been proposed to improve the navigation models. 
	However,  they are usually data inefficient and require a large amount of training data. In order to train these deep models, we need to construct a sufficient number of 3D synthetic environments and annotate the object information, which is exceedingly expensive, time-consuming, and even infeasible in real-world applications. Furthermore, it is hard for the trained embodied agent to transfer to different environments.
	
	It is worth noticing that when humans encounter a new task, they can quickly learn to solve it by transferring the meta-skills learned in a wide variety of tasks throughout their lives. This stands in stark contrast with the current deep reinforcement learning-based navigation methods, where the policy networks are learned from scratch. Instead, humans have an inherent ability to transfer knowledge across tasks and cross-utilize their knowledge, which offloads the burden of a large number of training samples. 
	
	\begin{figure*}
		\begin{center}
			\includegraphics[width=\textwidth]{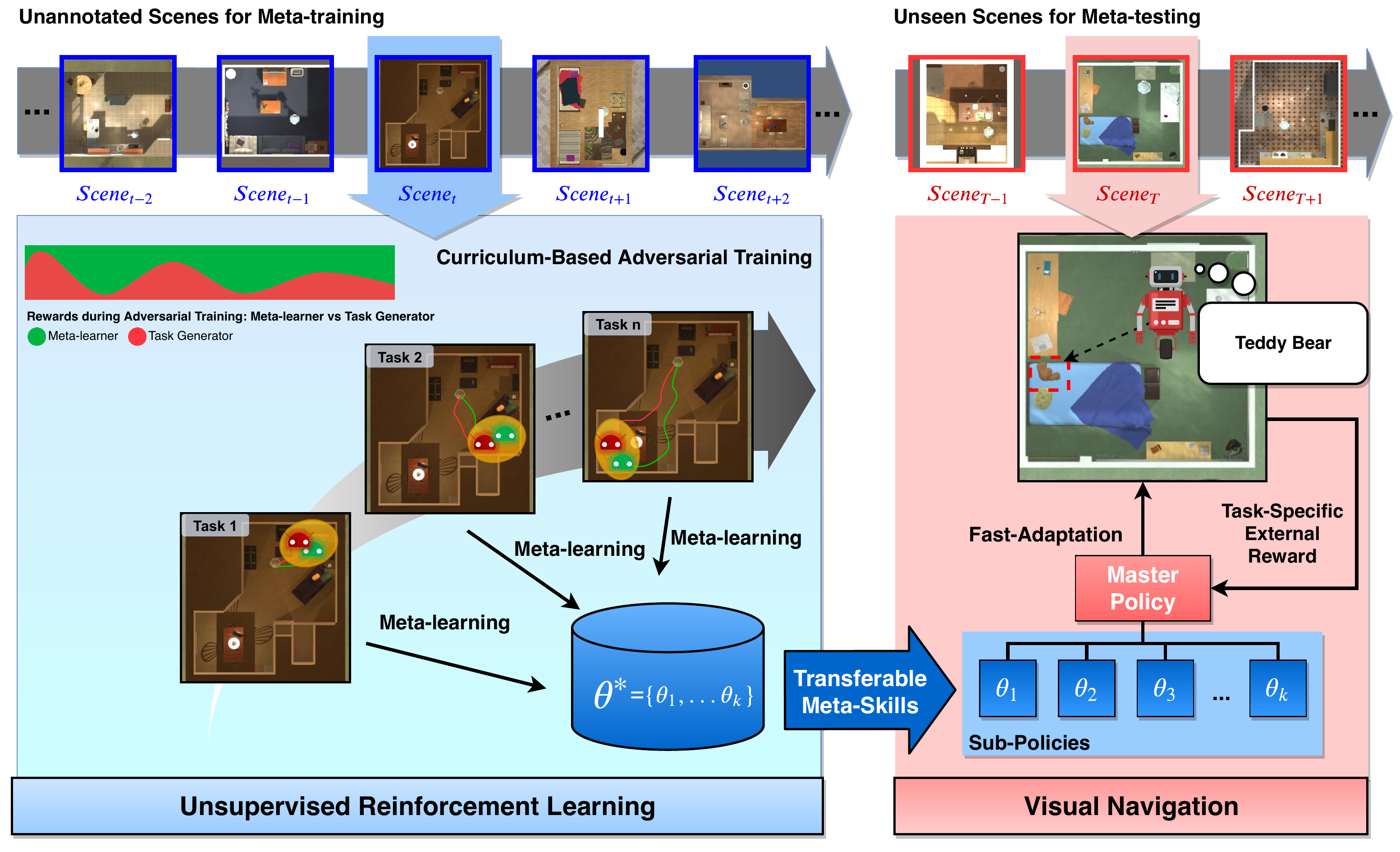}
		\end{center}
		\vspace{-0.5cm}
		\caption{\textbf{Overview of our ULTRA framework. } The blue part on the left is our adversarial training process, where the task generator automatically proposes a curriculum of increasingly challenging tasks, and the meta-learner learns to complete them. From these tasks, the meta-learner learns a set of transferable sub-policies. Then, on the right part, the meta-learner can fast adapt to visual navigation by just learning a new master policy, given the task-specific external reward. The $\theta_k$ is corresponding to the parameters of the $k$-$th$ sub-policy.}
		\label{overview}
		\vspace{-0.5cm}
	\end{figure*}

	Inspired by this fact, we seek the help of both meta-learning~\cite{nichol2018first, finn2017model}\ that learn quickly using a small amount of data and transfer learning~\cite{torrey2010transfer, weiss2016survey} that accelerate learning a new task through transferring knowledge from a related task that is already learned. In our work, we frame low-resource visual navigation as a meta-learning problem. At the meta-training phase, the environments are not annotated with object information, and we assume access to a set of tasks that we refer to as the meta-training tasks. From these tasks, the embodied agent (we call it as meta-learner) then learns a set of transferable sub-policies, each of which corresponds to a specific meta-skill (also called as motor primitives, \eg, bypass obstacles, go straight) by performing a sequence of primitive actions. 
	At the meta-testing phase, a few annotated environments with hand-specified rewards for visual navigation are provided. As illustrated in Figure \ref{overview}, after learning transferable sub-policies from meta-training scenes, the agent is solely required to learn a new master policy to combine the sub-policies such that it can fast adapt to visual navigation. During meta-training, the master policy is task-specific, and the sub-policies are shared for all tasks across scenes. The master policy determines the execution order of the sub-policies and is optimized to fast adapt to each meta-training task. The sub-policies are optimized for performance across tasks using gradient-based meta-learning algorithms~\cite{nichol2018first, finn2017model}. The hierarchical architecture~\cite{frans2017meta, sutton1999between, bacon2017option, florensa2017stochastic} that separates the entire policy into the task-specific part and task-agnostic part can also avoid meta-overfitting: typical gradient-based meta-learning algorithms can easily result in overfitting since the entire network is updated on just a few samples. 
	
	However, typical meta-learning methods~\cite{nichol2018first, finn2017model} require a sufficient number of hand-designed tasks for meta-training, which is not practical for an embodied agent. In this paper, we then propose a novel unsupervised reinforcement learning approach that automatically generate a curriculum of tasks without manual task definition. In our Unsupervised reinforcement Learning of TRAnsferable meta-skills (ULTRA) framework, the agent can efficiently learn transferable meta-skills and thus fast adapt to the new task by leveraging the meta-skills when entering a new environment. The main body of the framework is what we call \textsl{the curriculum-based adversarial training process}, where one agent (\textsl{task generator}) generates a curriculum of tasks with increasing difficulty. 
	The other agent (\textsl{meta-learner}) learns the meta-skills by accomplishing the generated tasks. After this unsupervised adversarial training process, the meta-learner can fast adapt to the new visual navigation task by just learning a new master policy to combine the learned meta-skills.
	
	Our experimental results show that our method significantly outperform the baselines by a large margin, and further ablation study demonstrates the effectiveness of each component. Additionally, qualitative analysis demonstrates the consistent behavior of the sub-policies. 
	In summary, our contributions are mainly four-fold:
	\vspace{-0.2cm}
	\begin{itemize}
		\item We propose a novel ULTRA framework to learn meta-skills via unsupervised reinforcement learning.
		
		\item The hierarchical policy of meta-learner separates the entire policy into the task-specific part and task-agnostic part, which reduces the probability of meta-overfitting and promises a faster convergence.
		
		\item Instead of manually designing tasks, we propose a novel curriculum-based adversarial training strategy, where the task generator automatically proposes increasingly difficult tasks to the meta-learner. Further, we define a diversity measure to encourage the task generator to generate more diverse tasks.
		
		\item We perform our experiments in low-resource setting, and experimental results show that our method significantly outperforms the baseline by $53.34\%$ relatively on SPL and requires only one-third number of iterations to converge, compared with the baseline.
	\end{itemize}

	\section{Related Work}
	\noindent
	\textbf{Visual Navigation. }Traditional navigation methods~\cite{blosch2010vision, cummins2007probabilistic, kidono2002autonomous, konolige2008outdoor, matthies1987error, thrun1998learning} typically employ geometric reasoning on a given occupancy map of the environment. They perform path planning~\cite{canny1988complexity, kavraki1996probabilistic, lavalle2006planning}to decide which actions the robot performs. Recently, many deep reinforcement learning (DRL) approaches~\cite{wortsman2019learning, mousavian2018visual, savva2017minos, wu2018building, yu2018guided,gupta2017cognitive, mirowski2016learning, zhu2017target}  have been proposed. 
	While these methods achieve great improvement, it is difficult to apply them to real-world situations since these DRL methods require a large number of training episodes and annotated environment information, which is time-consuming and exceedingly expensive. In our work, we focus on developing an unsupervised reinforcement learning method in the low-resource setting.
	
	\noindent
	\textbf{Meta-Learning. }Meta-learning, also known as learning to learn, optimizes for the ability to learn new tasks quickly and efficiently, using experience from learning multiple tasks. There are three common types of methods:\ 1) \emph{metric-based} methods~\cite{snell2017prototypical, sung2018learning, vinyals2016matching} that learn an efficient distance metric; 2) \emph{memory-based} methods~\cite{mishra2017simple, munkhdalai2017meta, oreshkin2018tadam, santoro2016meta} that learn to store experience using external or internal memory; and 3) \emph{gradient-based} methods~\cite{nichol2018first, finn2017model, hochreiter2001learning, ravi2016optimization, frans2017meta} model parameters explicitly for fast learning. Our method relies on a gradient-based meta-learning algorithm called Reptile~\cite{nichol2018first}.  The Reptile algorithm is aimed to learn a good parameter initialization during the meta-training process, where a large number of related tasks are provided. Thus, in the meta-testing process, the model can achieve good performance on new tasks after only a few gradient updates. An important difference is that our method does not require a large number of hand-designed tasks at the meta-training stage. 
	
	\noindent
	\textbf{Intrinsic Motivation-Based Exploration. }Intrinsic motivation or curiosity called by psychologists have been widely used to train an agent to explore the environment and create environment priors without external supervision. There are mainly two categories of intrinsic reward: 1) incentivize the agent to explore ``novel'' states~\cite{eysenbach2018diversity, gupta2018unsupervised, sukhbaatar2017intrinsic}; and 2) incentivize the agent to perform actions that reduce its predictive uncertainty of the environment~\cite{pathak2017curiosity}. 
	
	Sukhbaatar \textsl{et al.}~\cite{sukhbaatar2017intrinsic} introduce an adversarial training approach to unsupervised exploration, where one model proposes tasks and the other learns to complete it. In their work, the model for completing the tasks shares the whole parameters during training, and use the parameters as initialization for the downstream task. However, our work differs as we treat the adversarial training process as a sequence of independent meta-training tasks, and each task holds independent task-specific parameters. Also, there is no communication between two agents, whereas, in our work, the generator sends the target observation to the meta-learner, which contains the task information.
	
	Gupta \textsl{et al.}~\cite{gupta2018unsupervised} propose an unsupervised meta-learning method based on a recently proposed unsupervised exploration technique~\cite{eysenbach2018diversity}. They use the heuristic method to define intrinsic reward (\textsl{i.e. random discriminator, entropy-based method}), which automates the task generation process during meta-training. 
	Our work instead introduces an adversarial training strategy, which is more interpretable and efficient.
	
	\section{Method}        
	In this section, we first define the meta-learning setting for visual navigation. We then describe our ULTRA framework. Finally, we discuss how to transfer the meta-skills to visual navigation.
	
	\begin{figure*}
		\begin{center}
			\includegraphics[width=\textwidth]{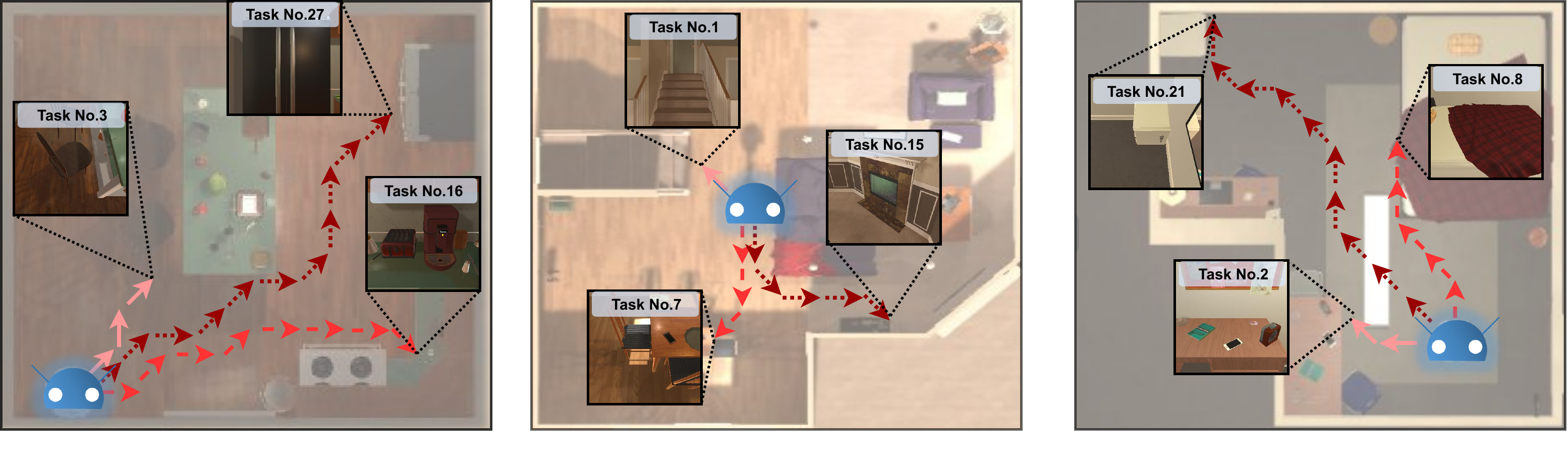}
		\end{center}
		\vspace{-0.7cm}
		\caption{\textbf{Graphical illustration of the task generator.} The generator starts from the same location~(denoted by the blue robot icon) and generates tasks for the meta-training. The level of difficulty~(represented by the darkness of the path) increases along the training process.}
		\label{qualitative}
		\vspace{-0.5cm}
	\end{figure*}
	
	\subsection{Problem Setup}
	Our goal is to learn meta-skills in an unsupervised manner and then transfer the acquired meta-skills to new tasks (\ie \textsl{visual navigation}). As illustrated in Figure~\ref{overview}, our approach has two stages: 1) In the meta-training stage, the agent learns transferable meta-skills via unsupervised reinforcement learning without human-specified reward functions. We use the curriculum-based adversarial training strategy to automatically generate a curriculum of meta-training tasks. 2) In the meta-testing stage, the agent is required to fast transfer to visual navigation task by utilizing the learned meta-skills. Training in this stage is fully supervised, but only a few training data is available.
	
	Note that, the automatically generated meta-training tasks are different from visual navigation in the meta-testing stage. During meta-training, the learning target is to recover the x, y and viewing angle of the agent according to egocentric RGB observations and an image given by the task generator (called image-driven navigation). Different targets correspond to different tasks. While during meta-testing, the input to the agent is not an image, but a language command~(\textsl{e.g., microwave}). The agent is required to understand various language commands and navigate to the objects specified by the commands in unseen scenes (called semantic visual navigation). 
	
	\subsection{Overview}
	As described in Figure \ref{overview}, our ULTRA framework mainly consists of three components: curriculum-based adversarial training strategy, shared hierarchical policy, and meta-reinforcement learning. During the curriculum-based adversarial training process, the task generator automatically proposes a curriculum of tasks, and the meta-learner learns to complete these tasks. Specifically, the architecture of the meta-learner is the shared hierarchical policy, which contains a master policy and a set of sub-policies. At each master-timestep, the master policy first selects a sub-policy to be activated, and then the chosen sub-policy performs primitive actions. The master policy is learned from scratch for each task and encodes the task-specific information. The sub-policies are shared and encapsulate meta-skills that can be transferred across all tasks. For each task generated by the task generator, the meta-learner first reinitialize the master policy and learns to combine the sub-policies to complete the task. After adapting the master policy to the new task, the meta-reinforcement learning algorithm is applied to optimize the sub-policies to excellent performance across tasks.
	
	\subsection{Curriculum-Based  Adversarial Training} \label{s3.3}
	In this setting, we have two agents: a task generator and a meta-learner. During each iteration, the task generator starts at the initial state $s_0 $, performs a sequence of actions, and finally stops at state $s_T$. Then, it sends its egocentric observation at the final state $s_T$ to the meta-learner. Given the observation $o_T$ at final state $s_T$, the goal of the meta-learner is to reach $s_T$ from $s_0$, which we call as a task. We initialize the meta-learner at state $s_0$, let it learn on this task for multiple episodes, and compute the success rate $r$. After that, the task generator proposes a new task, and the meta-learner repeats the above process. 
	
	
	Our goal is to automatically generate a curriculum of diverse tasks, where we first start with an easy task and then gradually increase the task difficulty. The reward function of the task generator consists of three components: a final reward based on the success rate, an intermediate reward that penalizes the task generator for taking too many steps, and a diversity measure that measures the diversity of the tasks.
	
	\noindent
	\textbf{Success Rate:} We use the success rate of the meta-learner after multiple episodes to measure the difficulty of the task and give the generator a final reward. The final reward is defined as:
	\begin{equation}
	R_f = k * (1-r)
	\end{equation}
	where  $k$ is a scaling factor, and $r$ is the success rate.
	
	\noindent
	\textbf{Step Efficiency:} At each timestep, the task generator will receive a negative constant intermediate reward. We penalize the task generator for taking too many steps, which encourages it to generate the easiest task that the meta-learner can not complete. In the first few iterations, the task generator can propose tasks by performing a small number of steps. Then as the capabilities of the meta-learner increase, more steps will be taken to generate more difficult tasks\ (qualitative examples in Figure \ref{qualitative}).
	
	\noindent
	\textbf{Task Diversity:} In order to explore wider state spaces for our meta-learner to build a better visual and physical understanding of the environment, we add an additional item in the task generator's reward function to encourage it to generate more diverse tasks. Formally, let $\pi $ denote the current policy, and $\pi'$ denote a previous policy. The diversity measure $D$ can be written as:
	\begin{equation}
	D = \sum_{s_t \in \tau} \sum_{\pi' \in \Pi}  D_{KL}(\pi'(\cdot|s_t)||\pi(\cdot||s_t))
	\end{equation}
	where $\tau$ is the trajectory from the current episode, $\Pi$ is the set of prior polices. We save the previous policy corresponding to the last four episodes in the set $\Pi$. We use KL-divergence to measure the difference between the current policy and the previous policies. The task diversity is aimed to incentivize the task generator to generate more diverse tasks that cover a larger state space of the environment.
	
	Formally, the task generator's total reward $R_G$ can be written as:
	\begin{equation}
	R_G = k*(1 - r) - \lambda*n + \eta*\sum_{s_t \in \tau} \sum_{\pi' \in \Pi}  D_{KL}(\pi'(\cdot|s_t)||\pi(\cdot||s_t))
	\label{e3}
	\end{equation}
	where $\lambda$ and $\eta$ are weight hyper-parameters, and $n$ is the number of actions that the task  generator executes.
	
	For meta-learner, we use the shared hierarchical policy. We train it using actor-critic methods\cite{mnih2016asynchronous} with rewards function that incentivizes it to reach the target.
	
	\subsection{Shared Hierarchical Policy}
	The shared hierarchical policy decomposes long-term planning into two different time-scales. At master-timestep, the master policy chooses a specific sub-policy from a set of sub-policies and then gives control to the sub-policy. As in~\cite{frans2017meta}, the sub-policy executes fixed N timesteps primitive actions~(\textit{e.g. MoveAhead, RotateLeft}) before returning control back to the master policy. 
	
	Formally, let $\phi$ denote the parameters of the master policy, and $\theta = \{\theta_1, \theta_2, ..., \theta_K \}$ denote the parameters of the K sub-policies. $\phi$ is the task-specific parameters, that is learned from scratch for each task. $\theta$ is shared between all tasks and switched between by task-specific master policies. For each task generated by the task generator during the adversarial training process, $\phi$ is randomly initialized at first and then optimized to maximize the total reward over multiple episodes, given fixed shared parameters $\theta$. 
	
	After fine-tuning the task-specific parameters $\phi$ to the task (called warm-up period), we take a joint update period, where both $\theta$ and $\phi$ are updated. The task-specific $\phi$ is optimized towards the current task, but the shared $\theta$ is optimized to excellent performance across tasks using gradient-based meta-learning algorithms. The details are discussed in the Sec \ref{s3.5}.
	
	\subsection{Meta-Reinforcement Learning on the Proposed Tasks} \label{s3.5}
	\begin{algorithm}[!t]
		\caption{Unsupervised Reinforcement Learning}
		\begin{algorithmic}[1]
			\State randomly initialize $\theta, \phi, \mu$
			\State $\Pi \longleftarrow [\ ]$
			
			\While{ not converged}
			\State $s_0 \longleftarrow e_i.start\_state$
			\State collect rollout $\tau_i^G(s_0, s_1, ... , s_T)$ using $\pi_{\mu}^G$
			\State $s^* \longleftarrow s_T$
			\State $o^* \longleftarrow o_T$
			\State set task $\tau_i = SetTask(s_0, s^*, o^*)$
			\For{$w = 0, 1, ... W$ (warmup period)}
			\State collect rollout $\tau_i^w$ using $\pi_{\phi_i, \theta}^M$
			\State $\phi_i \longleftarrow \phi_i + \alpha\nabla_{\phi}J(\tau_i^w, \pi_{\phi_i, \theta}^M)$
			\EndFor
			
			\State $\tilde{\theta} = \theta$
			
			\For{$j=0, 1, ... J$ (joint update period)}
			\State collect rollout $\tau_i^j$ using $\pi_{\phi_i, \tilde{\theta}}^M$
			\State $\phi_i \longleftarrow \phi_i + \alpha\nabla_{\phi}J(\tau_i^j, \pi_{\phi_i, \tilde{\theta}}^M)$
			\State $\tilde{\theta} \longleftarrow \tilde{\theta} + \alpha\nabla_{\theta}J(\tau_i^j, \pi_{\phi_i, \tilde{\theta}}^M)$
			\EndFor
			
			\State $\theta \longleftarrow \theta+ \beta(\tilde{\theta} - \theta)$
			
			\State Evaluate $R_G$ as Eq \ref{e3} and update $\pi_{\mu}^G$
			
			\If {$len(\Pi) == 4$}
			\State $\Pi.pop(0)$
			\EndIf
			
			\State $\Pi.append(\mu)$
			
			\EndWhile
		\end{algorithmic}
		\label{A1}
	\end{algorithm}
	
	Inspired by meta-learning algorithms~\cite{nichol2018first, finn2017model, hochreiter2001learning, ravi2016optimization, frans2017meta} that leverage experience across many tasks to learn new tasks quickly and efficiently, our method automatically learns meta-skills from a curriculum of meta-training tasks.
	
	\noindent
	\textbf{Background on Gradient-Based Meta-Learning:} Our method is inspired by prior work on a first-order gradient-based meta-learning algorithm called Reptile~\cite{nichol2018first}. The Reptile algorithm is aimed to learn the initialization of a neural network model, which can fast adapt to a new task. The Reptile algorithm repeatedly samples a task, training on it, and moving the initialization towards the trained weights on that task.
	
	Formally, let $\theta$ denote the parameters of the network, $\tau$ denote a sampled task, corresponding to loss $L_{\tau}$, and $\tilde{\theta}$ denote the updated parameters after $K$ steps of gradient descent on $L_{\tau}$. The update rule of the Reptile algorithm is as follows:
	\begin{equation}
	\theta \longleftarrow \theta+ \beta(\tilde{\theta} - \theta)
	\end{equation}
	where the ($\theta - \tilde{\theta}$) can be treated as a gradient that includes important terms from second-and-higher derivatives of $L_{\tau}$. Hence, the Reptile converges to a solution that is very different from the joint training.
	
	
	
	For Visual Navigation, our goal is for the agent to learn transferable meta-skills from the unsupervised adversarial training process. Therefore, we apply the Reptile algorithm to update the hierarchical police of the meta-learner. Different from the original Reptile algorithm that computes second-and-higher derivatives to update the whole parameters, we just apply it to update the parameters of the sub-policies and fix them during the test. Also, we treat ($\theta - \tilde{\theta}$) as a gradient and use SGD to update it.
	
	Algorithms \ref{A1} details our ULTRA that consists of four phases. Firstly, the task generator proposes a task. Secondly, the meta-learner joins in a warm-up period to fine-tune the master policy. Thirdly, the meta-learner takes a joint update period where both the master policy and sub-policies are updated. Finally, the task generator is updated based on the success rate of the meta-learner and repeats the above procedure.
	
	Formally, let $\pi_{\mu}^G$ denote the policy of the task generator parameterized by $\mu$, and $\pi_{\phi_i, \theta}^M$ denote the policy of the meta-learner parameterized by task-specific parameters $\phi_i$ and shared parameters $\theta = \{\theta_1, \theta_2, ..., \theta_K \}$. Firstly, we run the task generator and collect a trajectory $\tau_i^G(s_0, s_1, ... , s_T)$. We then set the task $\tau_i$ for the meta-learner by the initial state $s_0$, final state $s_T$, and the observation $o_T$ at the final state. Secondly, we initialize the meta-learner using the shared sub-policies and the random-initialized master policy. We then run a warmup period to fine-tune the master policy. More specifically, we run the meta-learner for $W$ episodes, and use the collected $W$ trajectories to update the master policy $\phi_i$ as follows:
	\begin{equation}
	\phi_i \longleftarrow \phi_i + \alpha\nabla_{\phi}J(\tau_i^w, \pi_{\phi_i, \theta}^M)
	\end{equation}
	where $J(\tau_i^w, \pi_{\phi_i, \theta}^M)$ is the objective function of any gradient-based reinforcement learning that uses the $w$-$th$ trajectory of task $\tau_i$ produced by policy $\pi_{\phi_i, \theta}^M$ to update the master policy $\phi_i$. In our work, we use Asynchronous Advantage Actor-Critic(A3C)~\cite{mnih2016asynchronous, wu2017scalable}.
	
	During the warmup period, the parameters of the shared sub-policies $\theta$ are fixed. After fine-tuning the master policy, we enter the joint update period, where we run the hierarchical policy for $J$ episodes, and update both $\phi_i$ and $\theta$ as follows:
	\begin{equation}
	\phi_i \longleftarrow \phi_i + \alpha\nabla_{\phi}J(\tau_i^j, \pi_{\phi_i, \tilde{\theta}}^M)
	\end{equation}
	\begin{equation}
	\tilde{\theta} \longleftarrow \tilde{\theta} + \alpha\nabla_{\theta}J(\tau_i^j, \pi_{\phi_i, \tilde{\theta}}^M)
	\end{equation}
	More specifically, we save the value of $\theta$ before the joint update period. After $J$ times iterations, we get the updated parameters $\tilde{\theta}$, and then we compute the gradient ($\theta - \tilde{\theta}$) and update the shared sub-policies $\theta$ using Reptile Algorithm. Finally, we compute the final reward of the task generator based on the success rate $r$, step efficiency, and the diversity.
	
	\subsection{Transferring to Semantic Visual Navigation}
	In the meta-testing stage, we fix the learned sub-policies from meta-training process, and employ the Asynchronous Advantage Actor-Critic(A3C)~\cite{mnih2016asynchronous, wu2017scalable} to train a new master policy on a few new scenes. The inputs of the master policy are the egocentric observation of current state and the word embedding of the target object~(\textsl{e.g., microwave}). In this stage, human-specified reward functions for visual navigation are available. If the agent reaches the target object within a certain number of steps, the agent receives a positive final reward. Also, it receives a negative intermediate reward at each step. Finally, we evaluate the performance on unseen scenes.
	
	\vspace{-0.3cm}
	\section{Experiments}
	In our experiments, we aim to (1) evaluate whether the agent can quickly transfer to visual navigation by leveraging the transferable meta-skills, given only a few training data, (2) determine whether the ULTRA is efficient than other unsupervised RL-based methods~\cite{eysenbach2018diversity, gupta2018unsupervised, pathak2017curiosity}, (3) determine whether the hierarchical policy promises a better transfer,  and (4) gain insight into how our unsupervised ULTRA works. 
	
	\subsection{Experimental Setup}
	
	We evaluate our approach in AI2-THOR~\cite{ai2thor} simulated environment, which is a photo-realistic customizable environment for indoor scenes and contains 120 scenes covering four different room categories: kitchens, living rooms, bedrooms, and bathrooms. We choose 60 scenes for meta-training, and 60 scenes for meta-testing. For the 60 meta-testing scenes, we further divide them into three splits~(\ie 20 scenes for supervised training, 20 scenes for validation, and 20 scenes for testing). During meta-training, object information and hand-specified rewards for visual navigation are not accessible, and the agent performs unsupervised reinforcement learning to learn transferable meta-skills. During meta-testing, all models are fine-tuned or learned from scratch on the training set, and are finally evaluated on the testing set. we choose the same set of navigational target object classes as~\cite{wortsman2019learning}, and the training reward is specific since the human-annotated labels are available. The action set $A$ consists of six unique actions (\textit{e.g. MoveAhead, RotateLeft, RotateRight, LookDown, LookUp, Done}).

	\begin{table}[t]
		\resizebox{\linewidth}{!}{
			\begin{threeparttable}
				\begin{tabular}{ lcccccc}
					\toprule
					&\multicolumn{2}{c}{All}  &\multicolumn{2}{c}{$L\geq5$}     \\
					&Success &SPL &Success &SPL\\
					
					\midrule
					Random                             &8.21                    &3.74                   &0.24      &0.09   \\
					A3C (learn from scratch)     &19.20                 &7.48                  &9.43      &4.13  \\
					DIAYN                                &17.23                 &6.30                  &8.72       &3.79      \\
					Curiosity                            &21.07                    &8.51                &10.31      &4.37       \\
					\textbf{Ours} - ULTRA                                 &\textbf{27.74}    & \textbf{11.47}   &\textbf{20.57}  &\textbf{8.04} \\
					\bottomrule
				\end{tabular}
			\end{threeparttable}
		}
		\caption{\textbf{Quantitative results.} We compare our approach with the baselines on testing data. Additionally, we report the results on trajectories where the optimal path length is at least 5 ($L\geq5$). Our ULTRA significantly outperforms the baselines, especially on $L\geq5$, indicating the superiority of our method on long-term planning.}
		\label{t1}
		\vspace{-0.4cm}
	\end{table}

	\noindent
	\textbf{Task and Evaluation Metric: }We use the averaged rewards on evaluation tasks during the training process to evaluate the learning speed, success rate to evaluate the navigation performance, and the success weighted by Path Length (SPL)\footnote{The SPL is defined as $\frac{1}{N} \sum_{i=1}^{N} S_i\frac{l_i}{max(p_i, l_i)}$, where $N$ is the number of episodes, $S_i$ is a binary indicator of success in episode $i$, $l_i$ is the shortest path distance, and $p_i$ is the path length. }~\cite{anderson2018evaluation} to evaluate the navigation efficiency. As~\cite{wortsman2019learning}, we report the performance both on all trajectories and trajectories, where the optimal path length is at least 5 ($L\geq5$).
	
	\begin{figure}[!t]
		\centering
		\includegraphics[width=\linewidth]{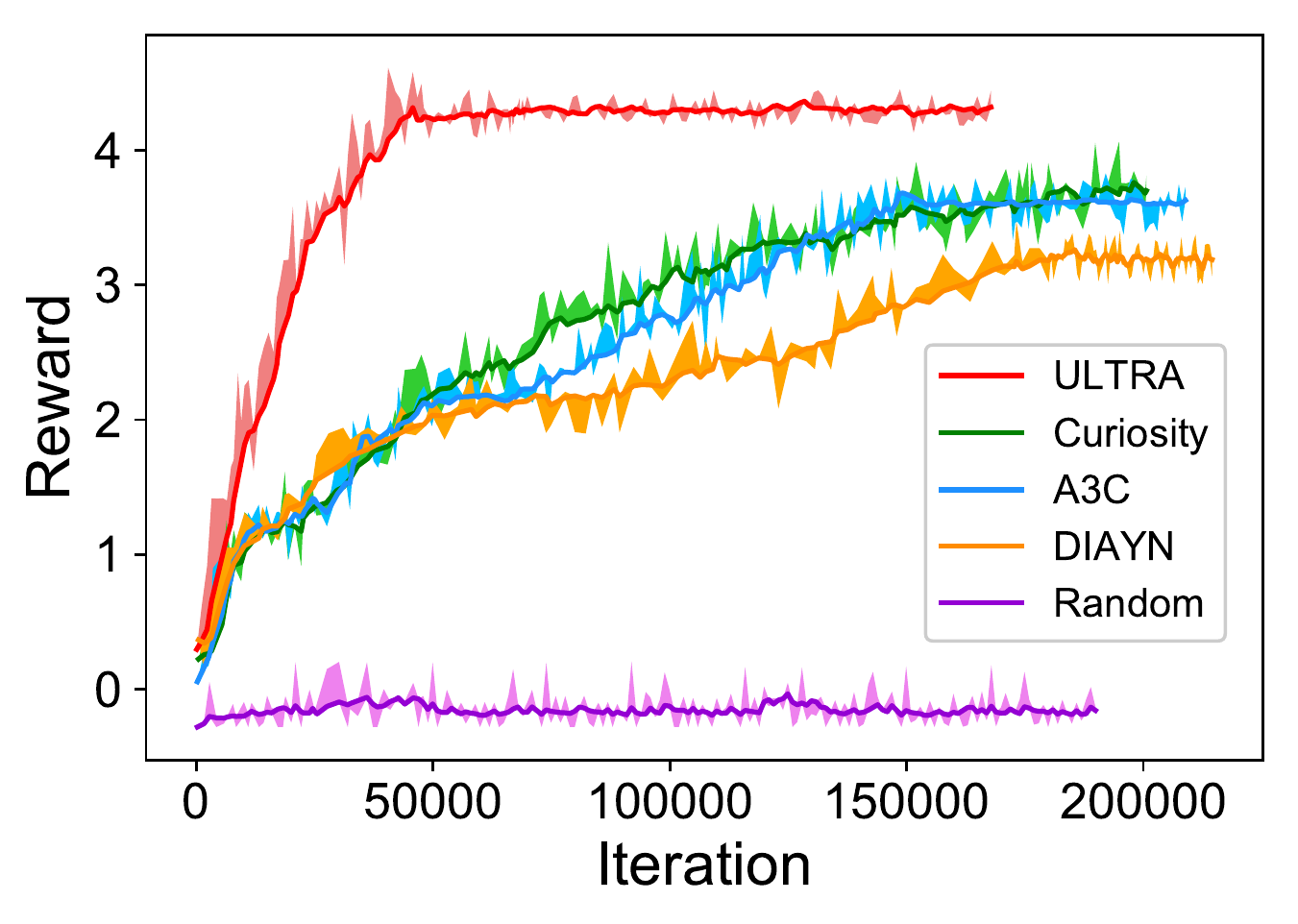}
		\vspace{-0.6cm}
		\caption{\textbf{Learning curves.} We report the rewards averaged across 10 evaluation tasks during meta-testing. }
		\label{results}
		\vspace{-0.5cm}
	\end{figure}
	
	\noindent
	\textbf{Baselines:} We compare our method with the following baselines: (1)~\textbf{Random policy: }The agent randomly execute an action at each timestep; (2)~\textbf{A3C (learn from scratch): }The architecture is the same as ours. However, there is no ULTRA process, and the whole hierarchical policy is directly learned from scratch in the  meta-testing stage with visual-navigation-specified rewards.
	
	We also compare to the state-of-the-art unsupervised RL-based methods: (3)~\textbf{Curiosity:}~\cite{pathak2017curiosity} The agent learns skills motivated by a curiosity reward, which serves as an intrinsic reward and is the error in an agent's ability to predict the consequence of its own actions in a visual feature space learned by a self-supervised inverse dynamics model.  (4)~\textbf{DIAYN:}~\cite{eysenbach2018diversity, gupta2018unsupervised} Diversity-driven methods hypothesize there is a latent variable~(control different skills) behind the agent state distribution and maximize the mutual information between the latent variable and the agent state distribution. They then combine the unsupervised skill acquisition (via DIAYN) with MAML\cite{finn2017model}. They train a discriminator to predict the latent variable from the observed state. As our ULTRA, DIAYN and Curiosity first conduct unsupervised reinforcement learning on scenes for meta-training and then are fine-tuned on training scenes for meta-testing with visual-navigation-specified rewards.

	\subsection{Results}
	We summarize the results of our ULTRA and the baselines in Table \ref{t1}. Also, we report the rewards averaged across 10 evaluation tasks during meta-testing in Figure \ref{results}. We observe that our approach can fast adapt to visual navigation, significantly outperforming all baselines not only in learning speed but also in performance. The number of iterations required for convergence of our ULTRA is about one-third of the baselines. Furthermore, our approach achieves the best success rate and SPL, especially when the trajectory length $L\geq5$, indicating the superiority of our method on long-term planning. DIAYN breaks down for visual navigation, because the same state can be reached via different skills from different initial state, which causes the discriminator performs at chance. Also, compared with A3C~(learn from scratch), the curiosity method makes limited improvement. We argue that the reason for this phenomenon is due to the complexity and diversity of the visual navigation environment, whose state space is always larger than the previous tasks.

	\begin{table}[t]
		\resizebox{\linewidth}{!}{
			\begin{threeparttable}
				\begin{tabular}{ lcccccc}
					\toprule
					&\multicolumn{2}{c}{All}  &\multicolumn{2}{c}{$L\geq5$}     \\
					&Success &SPL &Success &SPL\\
					
					\midrule
					A3C (learn from scratch)            &19.20                 &7.48                  &9.43      &4.13  \\
					A3C + random generator            &19.73                 &7.12                  &9.31      &4.47   \\
					A3C + hand-crafted generator    &20.57                &8.04                  &10.26      &4.28   \\
					\midrule    
					ULTRA                                 &\textbf{27.74}    & \textbf{11.47}   &\textbf{20.57} &\textbf{8.04}\\
					\quad \quad -- hierarchical policy   &24.27                 &10.54              &14.13      &5.61\\
					\quad \quad -- meta-RL update      &23.57                 &11.03               &14.02      &5.49\\
					\quad \quad -- adversarial training &20.23                 &8.35                &10.04      &4.33\\
					
					\bottomrule
				\end{tabular}
			\end{threeparttable}
		}
		
		\caption{\textbf{Ablation results.} We compare variations of our method with the A3C baseline augmented with random generator and hand-crafted generator.}
		\label{t2}
		\vspace{-0.5cm}
	\end{table}
	
	\vspace{-0.1cm}
	\subsection{Ablation Study}
	\noindent
	
	
	
	\noindent
	\textbf{Effect of Individual Components: }We conduct an ablation study to illustrate the effect of each component in Table \ref{t2}. We start with the final ULTRA model and remove the hierarchical policy, meta-RL algorithm, and the adversarial training, respectively. Furthermore, we augment the baseline with pre-training on meta-training scenes with a random generator and hand-crafted generator~\cite{kulhanek2019vision}. The augmented A3C baselines are first pre-trained on scenes for meta-training with a random generator or hand-crafted generator and then fine-tuned on training scenes for meta-testing. The random generator samples random location as the target, while the hand-crafted generator first samples the initial state closer to the target, and gradually increases the distance between the initial state and the target state. 
	The augmented baselines regard different targets as different episodes under a unified task and employ A3C to learn a policy. The augmented baselines correspond to typical pre-training methods that pre-train on a source task~(image-driven navigation) and transfer to the target task~(semantic visual navigation) by fine-tuning the parameters.
	
	Augmenting the baseline with pre-training on image-driven navigation proposed by the random generator or hand-crafted generator, we notice that there is no significant improvement and the performance is worse than Curiosity. The result reveals that image-driven navigation can not directly benefit semantic visual navigation. 
	
	The variation of ours without hierarchical policy uses a typical LSTM-A3C policy that updates the entire network during adversarial meta-training. We notice that the success rate drops 3.47 points, and the SPL drops 0.93 points, indicating that updating the entire policy on a few training samples of each meta-training tasks results in poor transferability. We then verify the strength of the meta-RL algorithm. As the pre-training baselines, we regard different targets as different episodes of the same task and update the parameters iteratively. Evidently, the meta-RL update improves upon the baseline by considering different targets as different meta-training tasks and updating the sub-policies by Reptile. Furthermore, the results of the last row (sample random location as meta-training tasks during unsupervised reinforcement learning) validate the superiority of curriculum-based adversarial training.
	
	\begin{figure}[!t]
		\centering
		\includegraphics[width=\linewidth]{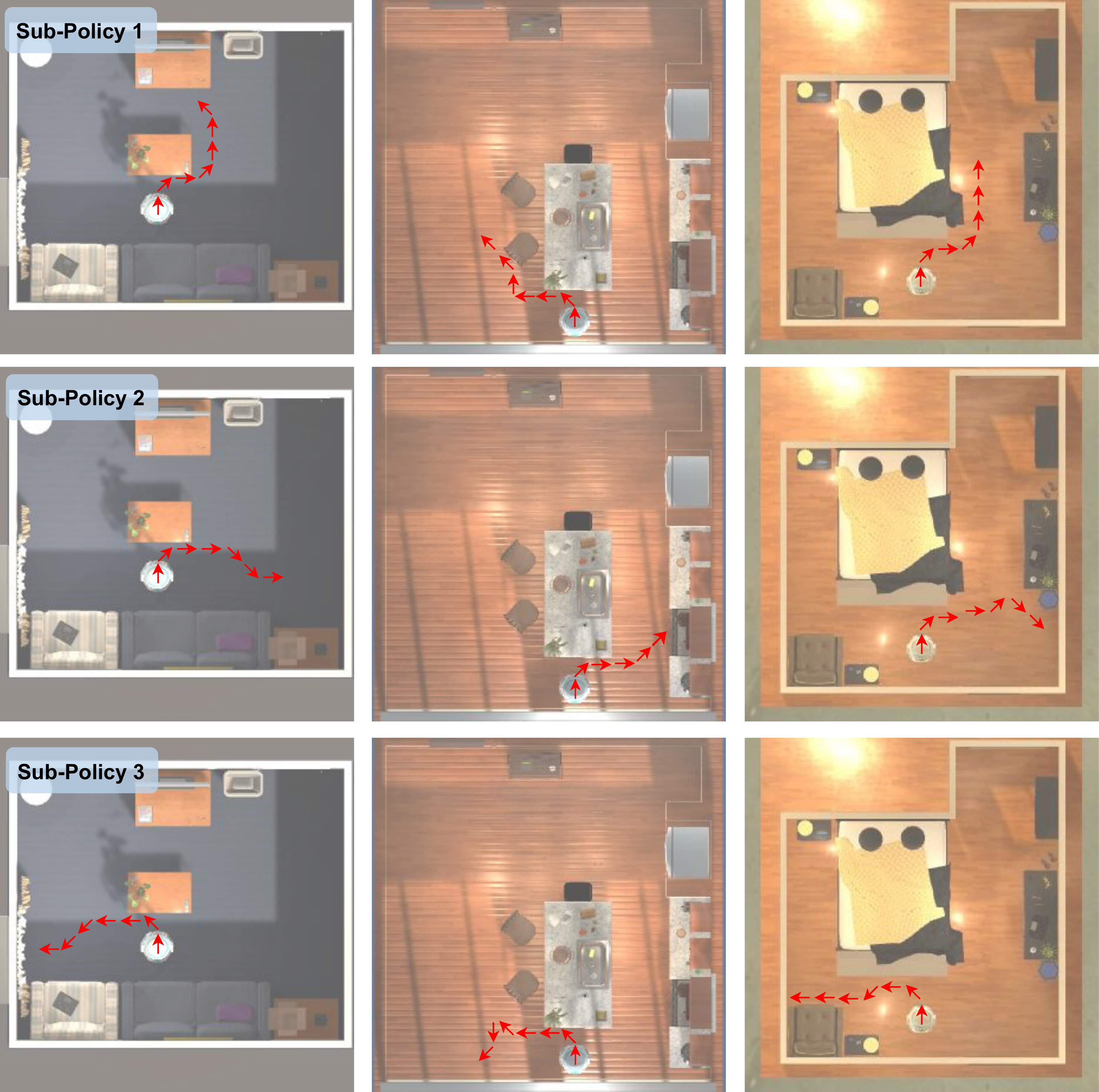}
		\vspace{-0.5cm}
		\caption{Visualization of the Sub-Policies.}
		\label{sub_vis}
		\vspace{-0.6cm}
	\end{figure}
	
\begin{figure}[!t]
		\centering
		\includegraphics[width=\linewidth]{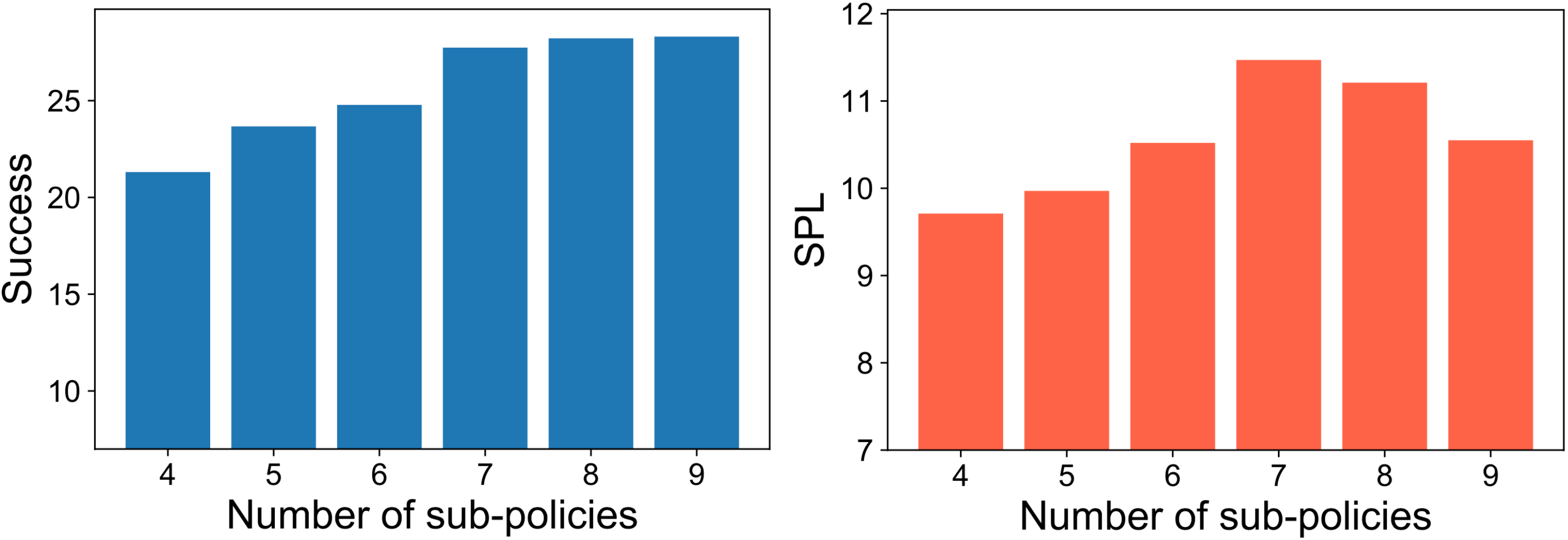}
		\vspace{-0.5cm}
		\caption{Ablation study of the number of the sub-policies.}
		\label{num_ablation}
		\vspace{-0.7cm}
	\end{figure}

	\noindent
	\textbf{Ablation of the Number of Sub-Policies: }To explore the impact of different numbers of the sub-policies, we modify the number of sub-policies. As illustrated in Figure \ref{num_ablation}, the success rate and SPL keeps increasing when the number of sub-policies is increased from 4 to 7. When we continue to increase the number of sub-policies, not only does the success rate not improve significantly, but SPL decreases because too many sub-policies results in confusion. In order to guarantee the performance and reduce the computational complexity, we set the number of the sub-policies to 7.
	
	\vspace{-0.1cm}
	\subsection{Qualitative Analysis}
	\noindent
	\textbf{Visualization of the task generator: }Figure \ref{qualitative} shows three qualitative examples. In each scenario, the tasks are generated starting from the same location. We can see that the difficulty of the tasks, corresponding to the length of the generated trajectories, increases as the serial number of the tasks goes up. Also, we can see that the generated trajectories in each scenario are in different directions, indicating that our task generator proposes diverse meta-training tasks.\\
	\noindent
	\textbf{Behavior of the Sub-Policy: }We execute sub-policies separately in different scenes to visualize the learned meta-skills. In Figure \ref{sub_vis}, trajectories shown in each row represent the same sub-policies initialized in different scenes, and trajectories shown in each column represent different sub-policies in the same location. As illustrated in Figure \ref{sub_vis}, the same sub-policy shows consistent behavior in different scenes. Sub-policy1 always bypasses obstacles and goes straight, sub-policy2 always turns right, and sub-policy3 always turns left. The consistency of the sub-policies demonstrates that our ULTRA has learned meaningful meta-skills.
	
	\vspace{-0.15cm}
	\section{Conclusions}
	In this paper, we introduce a novel ULTRA framework that enables the agent to learn transferable meta-skills in an unsupervised manner. Experiments show that our method can fast transfer to semantic visual navigation and outperform the baselines by a large margin. Additionally, we find that the sub-policies show consistent motor primitives. The ULTRA framework provides a new perspective that fuses the meta-learning and transfer-learning in a more interpretable way and in the future we plan to transfer the meta-skills to other tasks~(\textsl{i.e. Vision-and-Language Navigation~\cite{anderson2018vision}, Embodied Question Answering~\cite{das2018embodied} etc.}).
	
	\vspace{-0.2cm}
	\section*{Acknowledgment}
	This work has been supported in part by National Key Research and Development Program of China (SQ2018AAA010010), NSFC (No.61751209, U1611461), Hikvision-Zhejiang University Joint Research Center, Zhejiang University-Tongdun Technology Joint Laboratory of Artificial Intelligence, Zhejiang University iFLYTEK Joint Research Center, Chinese Knowledge Center of Engineering Science and Technology (CKCEST), Engineering Research Center of Digital Library, Ministry of Education. The authors from UCSB are not supported by any of the projects above.
	
	{\small
		\bibliographystyle{ieee_fullname}
		\bibliography{egpaper_final}
	}
	
\end{document}